\documentclass[runningheads]{llncs}
\usepackage[T1]{fontenc}
\usepackage{orcidlink}
\PassOptionsToPackage{hyphens}{url}
\usepackage{hyperref}
\usepackage{graphicx}
\begin{document}
\title{Unsupervised Learning for Industrial Defect Detection: A Case Study on Shearographic Data
    	\thanks{This preprint has not undergone peer review or any post-submission improvements or corrections. 
    		The Version of Record of this contribution is published in:
    		Artificial Intelligence XLII. Lecture Notes in Computer Science.
    		Springer Cham, 2026. DOI: \url{https://doi.org/10.1007/978-3-032-11442-6_22}}}
\titlerunning{Unsupervised Learning for Industrial Defect Detection}
\author{Jessica Plassmann\inst{1,2}\orcidlink{0009-0001-2257-457X} \and
Nicolas Schuler\inst{1,3}\orcidlink{0009-0007-4098-1244} \and
Georg von Freymann\inst{2,4}\orcidlink{0000-0003-2389-5532} \and Michael Schuth\inst{1}\orcidlink{0009-0007-7941-6808}}
\authorrunning{J. Plassmann \textit{et al.}}
\institute{Trier, University of Applied Science, 54293 Trier, Germany \\
\email{\{plassmaj, schulern, schuth\}@hochschule-trier.de} \and
Department of Physics and research center OPTIMAS, RPTU University Kaiserslautern-Landau, 67663 Kaiserslautern, Germany,\\
\email{georg.freymann@rptu.de} \and
University of Luxembourg, 4365 Esch-Belval Esch-sur-Alzette, Luxembourg \and
Fraunhofer Institute for Industrial Mathematics ITWM, 67663 Kaiserslautern, Germany}
\maketitle
\begin{abstract}
Shearography is a non-destructive testing method for detecting subsurface defects, offering high sensitivity and full-field inspection capabilities. However, its industrial adoption remains limited due to the need for expert interpretation. To reduce reliance on labeled data and manual evaluation, this study explores unsupervised learning methods for automated anomaly detection in shearographic images. Three architectures are evaluated: a fully connected autoencoder, a convolutional autoencoder, and a student-teacher feature matching model. All models are trained solely on defect-free data.
A controlled dataset was developed using a custom specimen with reproducible defect patterns, enabling systematic acquisition of shearographic measurements under both ideal and realistic deformation conditions. Two training subsets were defined: one containing only undistorted, defect-free samples, and one additionally including globally deformed, yet defect-free, data. The latter simulates practical inspection conditions by incorporating deformation-induced fringe patterns that may obscure localized anomalies. The models are evaluated in terms of binary classification and, for the student-teacher model, spatial defect localization.
Results show that the student-teacher approach achieves superior classification robustness and enables precise localization. Compared to the autoencoder-based models, it demonstrates improved separability of feature representations, as visualized through t-SNE embeddings. Additionally, a YOLOv8 model trained on labeled defect data serves as a reference to benchmark localization quality. This study underscores the potential of unsupervised deep learning for scalable, label-efficient shearographic inspection in industrial environments.

\end{abstract}
\section{Introduction}
Electronic Speckle-Pattern Shearing Interferometry (ESPSI), or shearography, is a well-established optical method for non-destructive testing (NDT), particularly effective for detecting subsurface defects in composites and other industrial components \cite{Tao:22a,Tao:22b}. Despite its high sensitivity to small surface deformations and rapid inspection capabilities, shearography is underused in serial applications due to the lack of fully automated evaluation processes \cite{Petry:21}.

A critical challenge in automating shearographic inspection is the reliance on supervised learning approaches, such as convolutional neural networks (CNNs) and object detection models like YOLO\cite{YOLO:16,Hussain:23}. These achieve their high performance primarily through large, extensive labeled datasets, such as ImageNet \cite{ImageNet:09}. Creating such datasets demands extensive manual effort and cost, and issues such as data scarcity, privacy, and labeling constraints further hinder dataset generation. Moreover, many companies possess vast unlabeled data unusable by supervised methods. \cite{Schmarje:21}

Studies like Li (2022) \cite{Li:22} have explored the use of synthetic data for generating datasets with sufficient and reproducible defects. However, generating synthetic data with realistic defect characteristics and variations remains challenging. Overly diverse synthetic datasets may lead to overfitting, particularly in large-scale applications \cite{Li:22}. This highlights the need to tailor datasets to specific components and testing conditions to ensure reliable defect detection. 

Given these challenges, unsupervised learning methods offer a promising alternative. Instead of relying on labeled defect data, these approaches learn the characteristics of defect-free components and identify deviations from this baseline \cite{Pang:21}. This method significantly reduces dependency on manually curated defect annotations. This paper explores the application of unsupervised learning for automating shearographic defect detection, with a particular focus on autoencoders \cite{Chen:23} and student-teacher models \cite{Wang:21}. These approaches are evaluated in terms of their effectiveness, robustness, and practical feasibility \cite{Wang:21b,Rudolph:22}. The comparison aims to determine the most suitable method for enabling reliable and automated shearographic inspection in industrial settings.

\section{Domain Background: Shearography}

Shearography is a non-destructive optical testing method for detecting subsurface defects in materials. The method measures deformation changes between two loading states with submicrometer sensitivity, enabling detection of minute surface displacement gradients. The component surface, which appears optically rough at the laser wavelength scale, is illuminated with coherent light, producing a granular speckle pattern due to random interference. This speckle pattern carries phase information and propagates through an interferometric setup to the camera. ~\cite{Schuth:17}

In the setup, the backscattered wavefront is laterally shifted (sheared) by tilting one beam path, causing interference between neighboring surface points. Each image pixel thus results from interference between spatially offset points on the object surface, enabling measurement of deformation gradients. ~\cite[p.~554]{Schuth:17}

\begin{figure}
    \centering
    \includegraphics[width=0.9\linewidth]{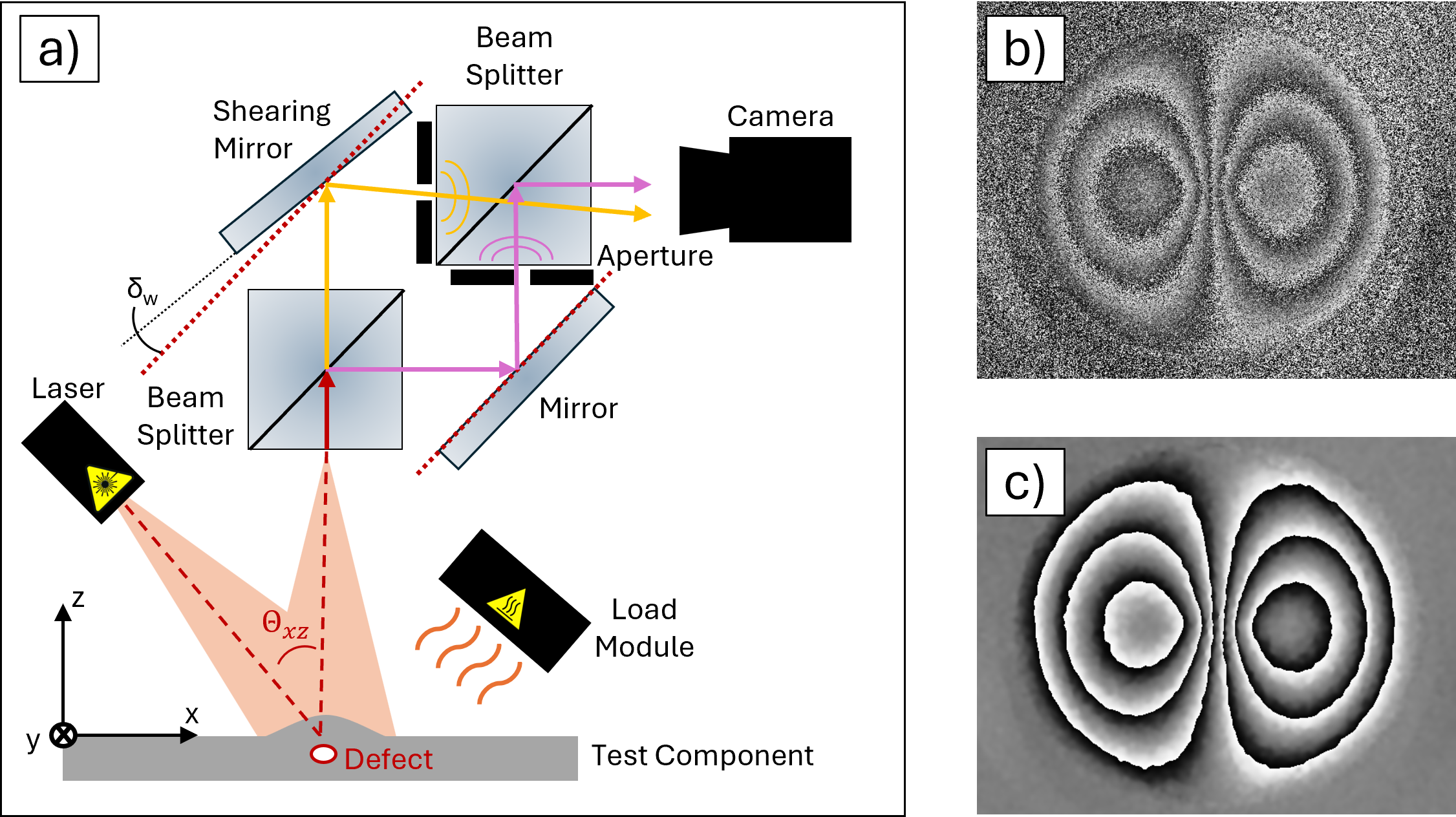}
    \caption{a) Shearography setup using a Mach–Zehnder interferometer with spatial phase shifting and thermal excitation, own illustration based on \cite[p.~97]{Petry:21}. b) Unfiltered and c) corresponding filtered phase image of a central out-of-plane deformation at a circular defect. \cite[p.~555]{Schuth:17}}
    \label{fig:setupshearo}
\end{figure}

Phase information necessary for quantitative evaluation is extracted using phase-shifting techniques. Temporal phase shifting relies on sequential image acquisition per loading state and thus requires static loading during the capture sequence. In contrast, spatial phase shifting enables single-frame evaluation, allowing measurements under dynamic loading conditions ~\cite{Petry:21}. Figure~\ref{fig:setupshearo}a illustrates the setup based on a Mach–Zehnder interferometer using spatial phase shifting and thermal excitation.

Measurements are based on comparing two states of the same object: a reference state and a state under external load. This load, typically thermal or mechanical, induces a deformation gradient that is captured as a differential phase map, known as a shearogram~\cite[pp. 314–327]{Schuth:17}. These maps visualize minute changes in surface deformation, enabling the detection of subsurface flaws through local anomalies in the phase pattern. However, environmental noise and speckle decorrelation can affect phase quality. Spatial or frequency-domain filtering is therefore applied to enhance relevant structures and suppress artifacts (Figure~\ref{fig:setupshearo}b–c)~\cite{Schuth:17}
. Although the technique originated in the 1970s, only recent advances in coherent light sources and computational processing have made its integration into industrial workflows viable \cite{Petry:21}.

\begin{figure}
    \centering
    \includegraphics[width=0.55\linewidth, trim=0 2.5cm 0 2.5cm, clip]{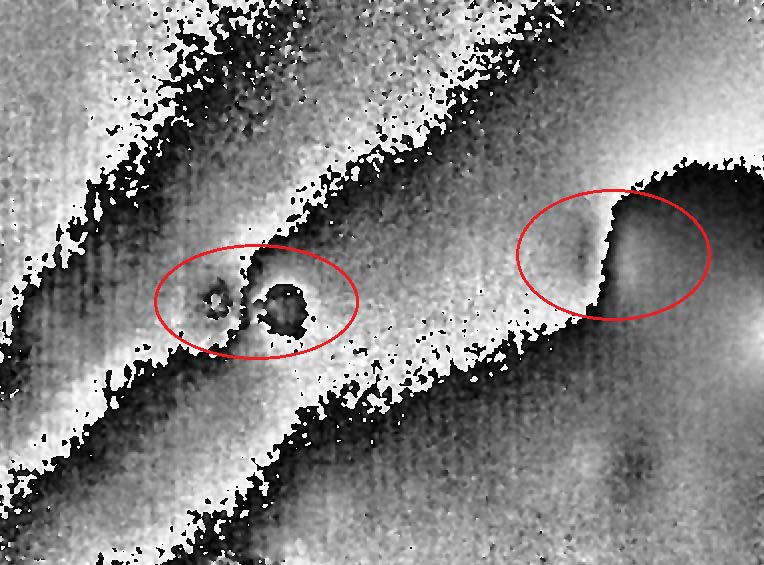}
    \caption{Filtered phase image with defect locations marked in red. Detection is challenging, especially when defect geometry is unknown and global deformations cause overlapping phase fringes. \cite[p.~556]{Schuth:17}}
    \label{fig:defectshearo}
\end{figure}

Despite its potential, shearography remains underutilized due to the difficulty of interpreting the resulting phase images. Phase fringes indicate deformation, but not necessarily defects. Accurate assessment requires prior knowledge of the expected deformation response under load. Subtle or localized anomalies caused by internal flaws can be obscured by various effects such as global deformation, measurement noise, or speckle decorrelation. As illustrated in Figure~\ref{fig:defectshearo}, even clearly defective regions may remain difficult to identify visually \cite{Schuth:17}.

As in many machine learning contexts, data availability is a key challenge. Industrial datasets are typically limited and imbalanced, with significantly more conforming parts than defective ones. Furthermore, defective samples with known flaws are rarely accessible due to confidentiality concerns. Consequently, this work explores unsupervised approaches trained exclusively on defect-free data to enable anomaly detection without labeled defects.

\section{Methodology}
The following section presents the architectural foundations of the implemented models, each designed to detect anomalies in shearographic image data. Two distinct approaches are introduced: an autoencoder-based reconstruction method and a Student-Teacher Feature Matching framework.

\subsection{Architecture}
\subsubsection{Autoencoder}
A widely used model for unsupervised anomaly detection is the autoencoder. It reconstructs input data by first encoding it into a lower-dimensional latent space and then decoding it back to its original form. For anomaly detection, the autoencoder is trained exclusively on defect-free samples, allowing it to learn the underlying distribution of normal data. When applied to defective samples, reconstruction fails due to deviations from the learned distribution, leading to a higher reconstruction error, which serves as an anomaly indicator. \cite{Chen:23}

Autoencoders offer flexibility in their architecture. A simple feed-forward autoencoder reduces input dimensions through multiple layers, while convolutional autoencoders use CNN-based encoders for feature extraction. The latter is particularly effective for image data, as it preserves spatial features. \cite{Chen:23}

Two autoencoder architectures were implemented. The first is a feed-forward autoencoder without convolutional layers, referred to as AE. The input image is resized to 96×50 pixels from an original resolution of 1920×1050 pixels and then flattened into a vector of size 4800. The encoder consists of four fully connected layers with output dimensions of 256, 128, 64, and 10, respectively. The latent representation of the data is thus a vector in $\mathbf{R}^{10}$. Each layer is followed by a dropout layer and a ReLU activation function to improve generalization and prevent overfitting.

The second implementation, referred to as ConvAE, is based on \cite{Gong:19a,Gong:19b}, an autoencoder originally designed for time-series anomaly detection, adapted for two-dimensional data. The CNN-based encoder consists of four convolutional layers with 96, 128, and 256 feature maps using 3×3 convolutional kernels. Each layer is followed by batch normalization and a LeakyReLU activation function. The decoder mirrors the encoder structure to reconstruct the input data from the latent representation.

\subsubsection{Student Teacher Model}
In this work, the Student-Teacher Feature Pyramid Matching (STFPM) approach \cite{Wang:21a,Wang:21b} was selected due to its high computational efficiency, widespread adoption, and competitive accuracy \cite{Myllaeri:23,Dutta:24}. The model has been successfully applied and adapted across various domains, making it a robust choice for detecting defects in shearographic inspections \cite{Schwarz:24a,Valjakka:23,Zhu:23,Yamada:21}.

\begin{figure}
    \centering
    \includegraphics[width=0.9\linewidth]{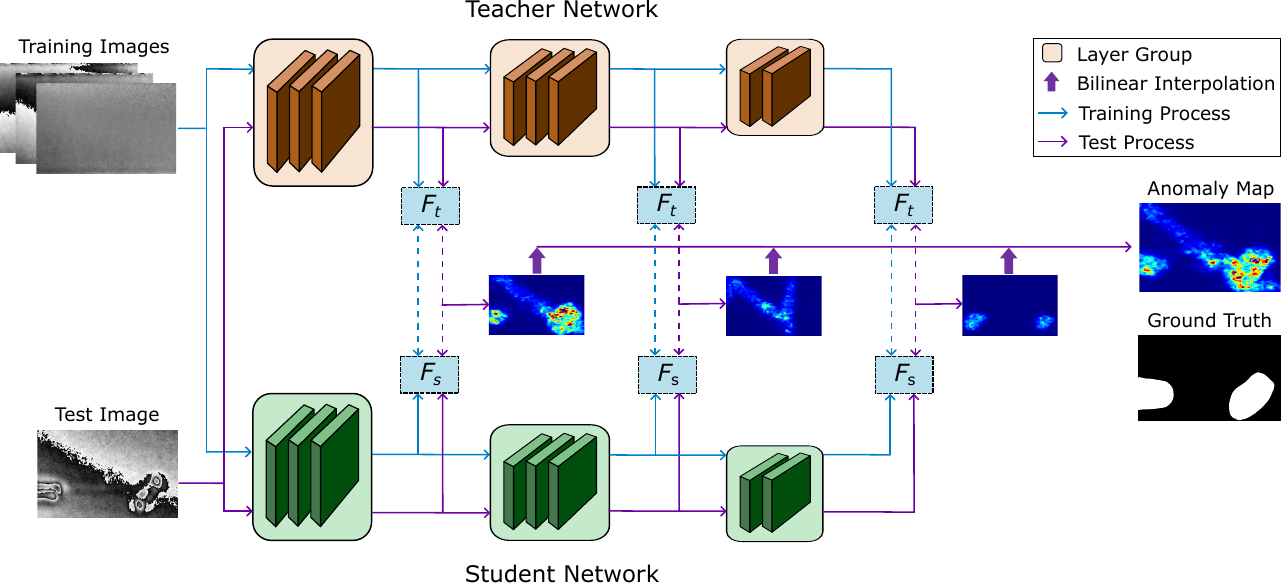}
    \caption{Architecture of the Student–Teacher Feature Pyramid Matching model, adapted from \cite{Wang:21}.}
    \label{fig:STFPMImp}
\end{figure}

The STFPM architecture, shown in Figure~\ref{fig:STFPMImp}, consists of two neural networks with identical structures, referred to as the Student and the Teacher. Unlike siamese networks, these architectures do not share trainable weights. The Teacher network is a pre-trained model that remains frozen during training, serving as a reference by providing feature representations of defect-free inputs. The Student network is trained exclusively on non-defective samples and learns to approximate the feature representations extracted by the Teacher. The training objective is to minimize the discrepancy between the feature outputs of both networks, with the loss function computed as the cumulative difference between the Student's and Teacher’s activations across multiple layers. \cite{Wang:21}

Once trained, the model produces a heatmap that visualizes the pixel-wise differences between the feature representations of the Student and the Teacher. Since the model is trained solely on non-defective data, it learns to reproduce normal feature distributions with minimal error. When applied to defective samples, however, the reconstruction error increases, particularly in the regions where anomalies are present. This characteristic enables STFPM not only to classify defective and non-defective samples but also to localize defects at a pixel level.

For the implementation in this work, the official STFPM model \cite{Wang:21a} was adopted and integrated into the existing class structures. Both the Student and Teacher networks use a ResNet18 architecture, with the pre-trained weights for the Teacher being sourced from the TorchVision model zoo \cite{PyTorch:}. In addition to the core implementation of STFPM, the model was embedded into multiple classes designed to facilitate visualization and analysis of the generated heatmaps. These include peak detection mechanisms to assist in the localization of potential defects within the input images.

Two variants of the student-teacher model were implemented: STFPM (Peaks) and STFPM (Means). Both share identical model weights; the only difference lies in the scoring strategy. STFPM (Peaks) uses the highest anomaly score from the predicted heatmap, whereas STFPM (Means) computes the average score over the input region.

\subsection{Dataset Preparation and training Setup}
A custom specimen was designed to simulate industrial conditions while ensuring reproducibility. A 500~mm × 50~mm acrylic glass strip was coated with a 1.2~mm thick, 45~mm wide rubber layer. Thin inserts of heat-shielding foil were embedded between the materials to simulate subsurface defects. These air-filled inclusions remain invisible from the surface but are verifiable from the transparent backside, providing labeled ground truth.

The specimen’s dimensions match the field of view of the shearography optics and the travel range of a UR5e robot arm, allowing automated scanning in 10~mm increments. Each frame contains at most two defects, enabling isolated evaluation. Data were acquired under two conditions: minimal global deformation through rigid fixation, and varied thermal excitation to induce diverse global deformation patterns. This allowed for capturing both localized defect responses and representative background variation.

The complete dataset comprises 10,498 shearography recordings acquired under controlled laboratory conditions. These recordings are divided into three categories. A total of 4,311 images contain deliberately introduced subsurface defects. Another 2,537 recordings represent defect-free specimens measured under fixed boundary conditions, which suppress global deformation. The remaining 3,650 images were acquired from defect-free specimens without fixation, leading to varying global deformation patterns and corresponding fringe structures.

The defective samples were manually annotated by a domain expert. Each annotated instance consists of a set of four temporally offset images capturing the same region under identical loading conditions, accompanied by bounding boxes marking the observed defect indications. The defect-free recordings serve two complementary roles: the measurements under fixed boundary conditions represent an idealized baseline free of global deformation, while the unfixed specimens introduce realistic variability in the form of fringe patterns caused by global deformation, thereby increasing the complexity of distinguishing local defect-related features from non-defect-induced surface behavior.

Based on the available dataset, two subsets were defined to address different evaluation scenarios in anomaly detection. The first, referred to as Subset A, includes only defect-free images acquired under fixed conditions, in which no global deformation patterns are present. This configuration yields a controlled baseline, consisting of 2,020 images for training, 254 for validation, and 254 for testing. Subset B includes all 6,187 defect-free recordings, incorporating both undistorted and globally deformed samples. In this case, the training set comprises 4,582 images, with 584 used for validation and 697 for testing. In both variants, defective samples are added to the validation and test sets to evaluate binary classification performance under realistic conditions. Specifically, 714 annotated defect samples are included in the test set and 391 in the validation set. These defective samples are excluded from anomaly detection training and reserved for model evaluation.

To quantitatively assess the localization performance of the proposed STFPM-based anomaly detection, a dedicated subset of the annotated defect data was used for training a YOLOv8 detection model \cite{yolov8}. The same 714 and 391 samples that were already included in the anomaly detection test and validation sets were reused to ensure consistency across evaluations. The remaining 3,523 annotated defective samples were reserved exclusively for training the YOLOv8 model, ensuring an unbiased benchmark.

Additionally, to evaluate the separability of the learned feature representations and the underlying structure of the data, t-distributed stochastic neighbor embedding (t-SNE) \cite{cai:2022} was applied.

All experiments were conducted on a Windows 11 desktop PC equipped with a 13th Gen Intel(R) Core(TM) i7-13700 processor, 128~GB RAM, and an NVIDIA RTX 4000 GPU with 16~GB VRAM. All models were implemented in Python 3.11 using the PyTorch framework~\cite{Pytorch}, with CUDA~\cite{cuda} support for accelerated training. Code and  dataset will be made available at~\cite{Git:Data}. 

\subsection{Model Comparison and Evaluation Metrics}

The evaluation of the proposed method addresses two key aspects: binary classification between defect-free and defective samples, and the quantitative assessment of localization performance. 

Classification performance was assessed using the receiver operating characteristic (ROC) curve~\cite{Hoo:17} and the precision-recall (PR) curve~\cite{Cook:20}. The area under the ROC curve (AUC) quantifies the trade-off between true positive rate and false positive rate, with an AUC of 0.5 indicating random performance for binary classification. The PR curve illustrates the relationship between precision and recall, with the average precision (AP) summarizing precision across all recall levels.

To quantify localization accuracy, the STFPM and YOLO models were evaluated using common metrics from object detection~\cite{Rez:19}. These include the intersection over union (IoU), the mean average precision (mAP), mAP at fixed IoU thresholds of 0.5 and 0.75 (mAP@0.5, mAP@0.75), and the mean average recall (mAR) at one and ten predictions per image (mAR@1, mAR@10). The IoU between a predicted region $B$ and a ground truth region $A$ is defined as the ratio of their intersection over their union. In cases with multiple predictions per ground truth region, only the prediction with the highest IoU is considered.

Since only a single defect class is evaluated, the mean average precision equals the average precision ($mAP = AP$). The fixed IoU thresholds define the criteria for correct matches, while the recall metrics quantify the fraction of correctly detected regions under limited prediction counts. All localization metrics were computed using the mean average precision implementation from the TorchMetrics library~\cite{Detlefsen:22}. As the Student-Teacher model does not produce conventional classification scores, the maximum anomaly score within each predicted region was used as a confidence proxy for ranking and evaluation.

\section{Results}

\subsection{Binary Classification}
\begin{figure}[htbp]
    \centering
    \includegraphics[width=\textwidth]{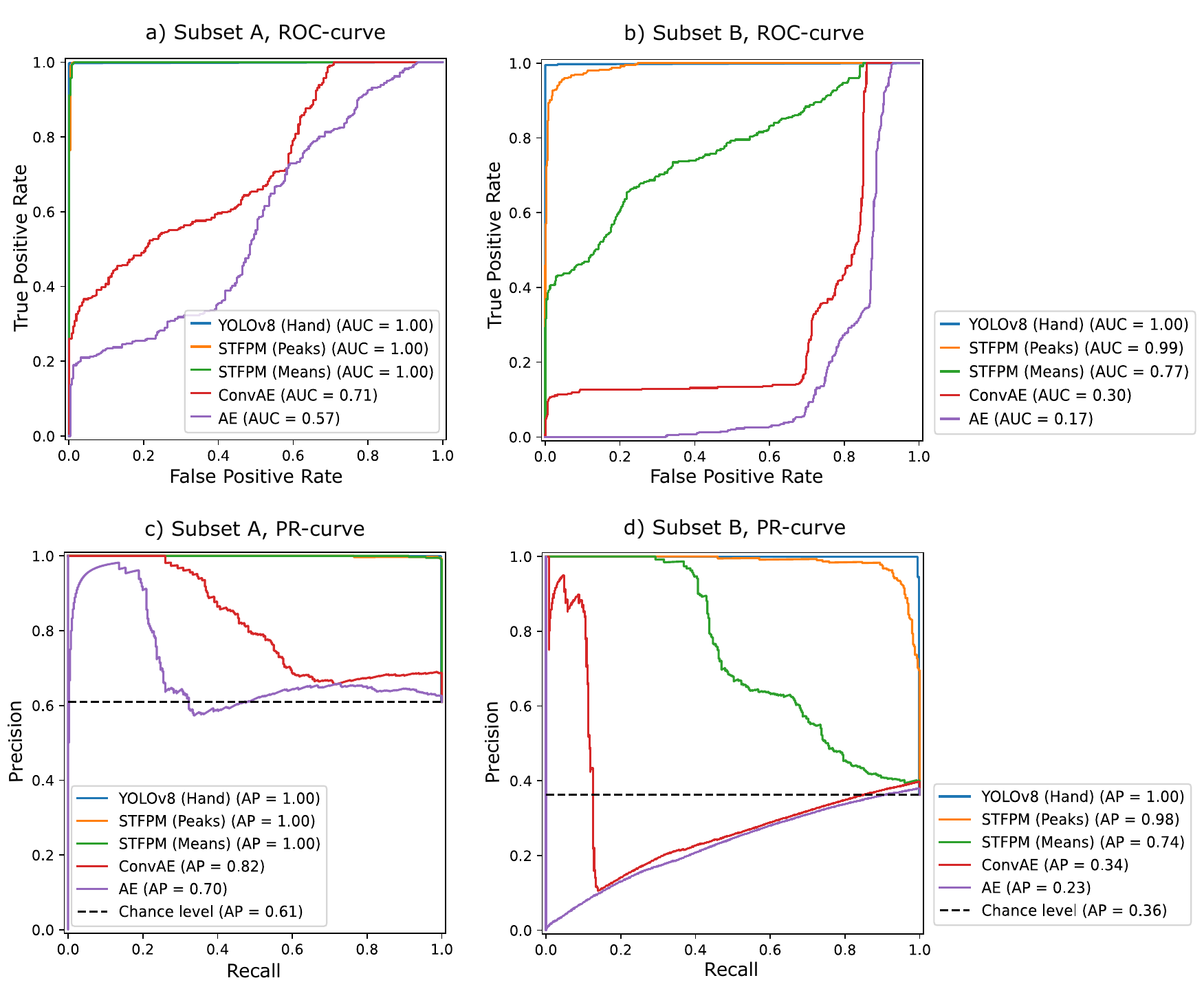}
    \caption{ROC and PR curves for the binary classification task on Subset~A and Subset~B. Each plot compares unsupervised models (STFPM Peaks/Means, ConvAE, AE) against the supervised YOLOv8 baseline. (a) ROC Subset~A, (b) ROC Subset~B, (c) PR Subset~A, (d) PR Subset~B.}
    \label{fig:roc_pr_curves}
\end{figure}

The corresponding ROC and PR curves for both subsets are shown in Figure~\ref{fig:roc_pr_curves}, illustrating the performance differences across all models. 
For Subset A, the STFPM-based methods achieve perfect performance, with AUC and AP values of 1.0, matching the supervised YOLOv8 baseline. The ConvAE model achieves an AUC of 0.71 and an AP of 0.82, while the simple Autoencoder performs notably worse (AUC of 0.57, AP of 0.70). Given a chance level of 0.61 for AP in this subset, only STFPM and ConvAE yield results above random performance.

In Subset B, the supervised YOLOv8 again achieves perfect scores (AUC and AP of 1.0). STFPM Peaks maintains strong performance with an AUC of 0.99 and an AP of 0.98, while STFPM Means shows reduced but still competitive results, with an AUC of 0.77 and an AP of 0.74. In contrast, ConvAE and AE exhibit significantly lower performance in this more challenging setting, achieving AUC values of 0.30 and 0.17, and AP values of 0.34 and 0.23, respectively. The chance level for AP in Subset B is 0.36.

\subsection{Dataset}
\begin{figure}[htbp!]
    \centering
    \includegraphics[width=\textwidth]{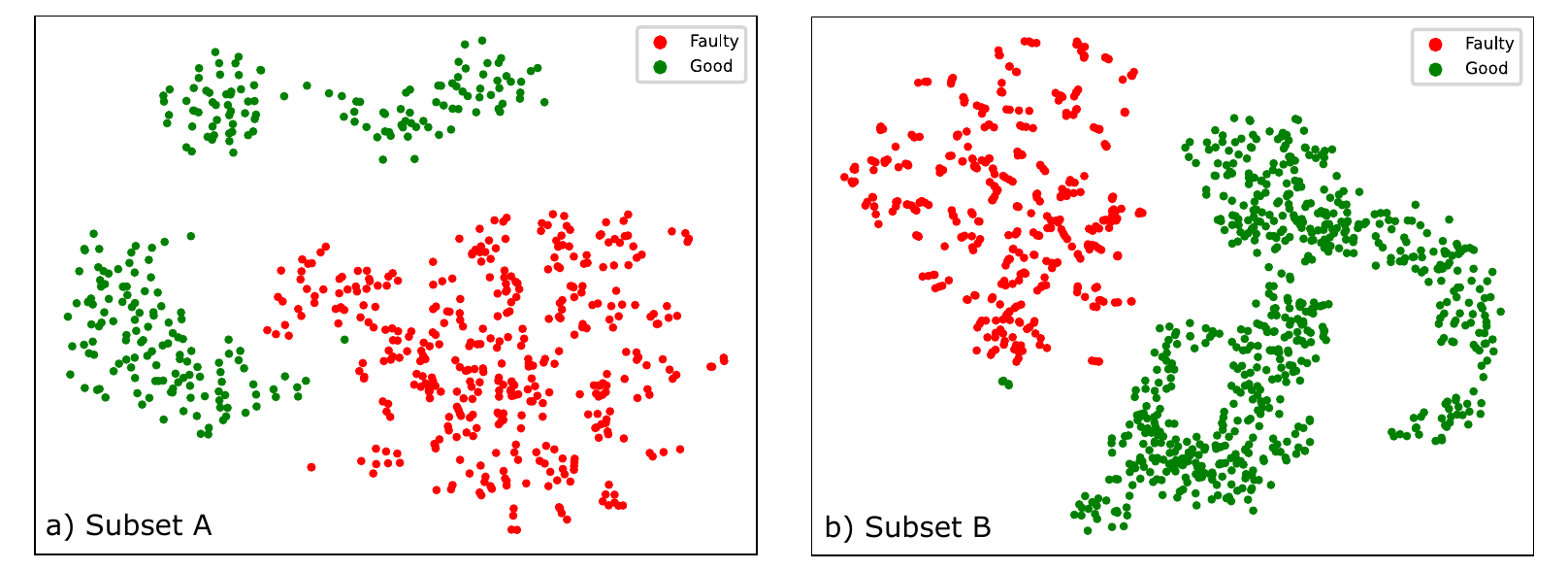}
    \caption{t-SNE embeddings illustrating the feature separability of defect-free and defective samples for (a) Subset~A, trained exclusively on undistorted data, and (b) Subset~B, which includes globally deformed defect-free samples.}
    \label{fig:tSNE}
\end{figure}

The t-SNE visualization in Figure~\ref{fig:tSNE} illustrates the learned feature space for both training subsets. Defect-free samples are shown in green, while defective samples are marked in red. In both subsets, a clear separation between classes is observed, indicating that the learned representations effectively capture discriminative characteristics.

In Subset A, which consists solely of undistorted defect-free samples, this separation is particularly pronounced, with only one defect-free sample located near the defective cluster. Furthermore, the defect-free samples in Subset A exhibit a distinct internal structure, forming at least two well-separated clusters. Subset B, containing globally deformed but non-defective samples, also demonstrates good overall class separability. However, two defect-free samples lie closer to the defective cluster boundary. Unlike Subset A, the defect-free samples in Subset B form a more contiguous cluster but contain noticeable low-density regions, suggesting non-uniform distribution within the normal class. These findings indicate that both subsets enable clear class separation while revealing meaningful intra-class variability within defect-free samples.

\subsection{Localization}
Table~\ref{tab:Loc} presents the localization results for the STFPM method, evaluated on Subset A and Subset B. For this experiment, only the Peaks variant was used. The predicted anomaly maps were smoothed and then binarized using fixed thresholds, which were optimized separately for each subset based on validation data. A threshold of 0.1 was selected for Subset A, and 0.001 for Subset B. The results are compared to those of a supervised YOLOv8 model trained with pixel-wise defect annotations.

\begin{table}[htbp]
\caption{Localization performance comparison of STFPM and YOLOv8.}
\label{tab:Loc}
\centering
\setlength{\tabcolsep}{3pt} 
\renewcommand{\arraystretch}{1.2} 
\begin{tabular}{lcccccc}
\hline
\textbf{Model} & \textbf{IoU} & \textbf{mAP} & \textbf{mAP@50} & \textbf{mAP@75} & \textbf{mAR@1} & \textbf{mAR@10} \\
\hline
YOLOv8              & 0.8695 & 0.7435 & 0.9901 & 0.9195 & 0.5817 & 0.7884 \\
STFPM (Subset A)    & 0.6965 & 0.3389 & 0.7421 & 0.1727 & 0.3209 & 0.4782 \\
STFPM (Subset B)    & 0.7113 & 0.3852 & 0.7832 & 0.2611 & 0.3703 & 0.5127 \\
\hline
\end{tabular}
\end{table}

\section{Discussion}
\subsection{Binary Classification}
The results indicate a clear divergence in performance between the STFPM-based methods and the autoencoder-based models across both subsets. For Subset A, representing near-ideal conditions, the STFPM Peaks and Means approaches achieve perfect classification performance, matching the supervised YOLOv8 baseline with AUC and AP scores of 1.0. This demonstrates that the STFPM methods effectively capture the discriminative features necessary to distinguish defective from defect-free samples under these conditions.

The STFPM Means method is conceptually similar to the autoencoder models, as it computes the average reconstruction error over the heatmap, analogous to the aggregation of the autoencoder’s reconstruction error. Nevertheless, STFPM Means consistently achieves higher performance than both the convolutional autoencoder and the simple autoencoder, which exhibit moderate and lower performance, respectively (ConvAE AUC 0.71, AP 0.82; AE AUC 0.57, AP 0.70). Furthermore, STFPM Peaks, which detects localized maximum errors instead of averages, attains the best performance among the unsupervised approaches, underscoring the benefit of capturing localized anomalies over global averages.

This pattern becomes even more pronounced on Subset B, which introduces global deformations and thus greater complexity. While YOLOv8 and STFPM Peaks maintain near perfect scores (AUC and AP of 1.0 and 0.99 / 0.98 respectively), STFPM Means experiences a reduction in performance (AUC 0.77 and AP 0.74) but still notably outperforms both autoencoder models, which perform below or near chance level (ConvAE AUC 0.30 and AP 0.34; AE AUC 0.17 and AP 0.23). These results further confirm the limitations of the autoencoder architectures in this setting.

Overall, these findings suggest that the relatively poor performance of the autoencoder models is likely attributable to architectural or training limitations rather than an inherent lack of class separability in the data. This is supported by the t-SNE visualization in Figure~\ref{fig:tSNE}, which reveals well-defined clusters corresponding to the two classes. The moderate success of the STFPM Means method further indicates that the data contain discriminative features accessible to appropriate models. To fully exploit this intrinsic separability, future work should investigate alternative model architectures or enhanced preprocessing techniques. In contrast, the consistently strong performance of the STFPM Peaks method underscores its robustness and effectiveness for anomaly detection, even under challenging conditions with significant global deformations.

\subsection{Localization}
Despite being an unsupervised approach, STFPM demonstrates promising localization capabilities across both datasets. Since localization was evaluated exclusively on defective samples, the YOLO model is unaffected by the subset definitions and can be compared consistently across both.
The localization performance depends strongly on the chosen thresholds. STFPM tends to generate multiple smaller bounding boxes for larger defect areas, which may be detected as separate anomaly regions instead of one unified defect. This fragmentation naturally affects metrics such as mean Average Precision and mean Average Recall, since these metrics penalize excessive or overlapping detections.

\begin{figure}[htbp]
    \centering
    \includegraphics[width=0.9\textwidth]{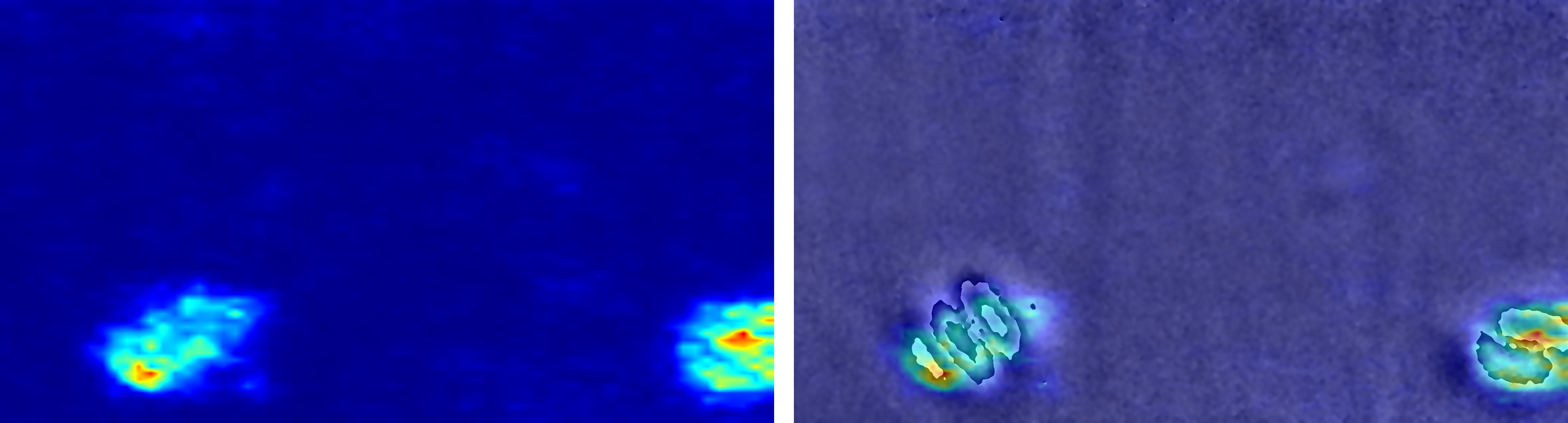}
    \caption{Exemplaric defect localization using STFPM. Left: predicted anomaly heatmap. Right: same heatmap overlaid on the original shearography image. High anomaly scores (red) correspond to localized defect regions.}
    \label{fig:localization_example}
\end{figure}

Despite the unsupervised nature of the method and the variability in defect size and appearance, overall detection quality remains high. Figure~\ref{fig:localization_example} illustrates this with a representative sample: the left image shows the anomaly heatmap generated by STFPM, while the right image presents the corresponding sample overlaid with the heatmap. The visualization highlights the model’s ability to localize defects effectively, although spatial accuracy may occasionally be limited by thresholding or peak fragmentation.
For illustration, Figure~\ref{fig:localization_example} shows an example image with a defect, the corresponding anomaly heatmap generated by STFPM, and the detected peak localization. This visualization highlights how STFPM successfully identifies the defect area despite some spatial imprecision.

\section{Conclusions}
This study demonstrates that STFPM, especially its Peaks variant, is a robust unsupervised approach for defect detection and localization in shearographic inspection of industrial components. It reliably distinguishes between defect-free and defective samples, even under complex conditions with global deformations, closely matching the performance of supervised YOLOv8 models that require extensive labeled data. In contrast, autoencoder-based models show limited effectiveness in this domain, likely due to architectural constraints rather than inherent data challenges, as confirmed by the clear class separability observed in feature space visualizations.

Importantly, STFPM's unsupervised nature addresses a key industrial challenge: enabling automated evaluation of shearographic data without costly, time-consuming defect annotations. While localization precision can be further improved, the promising results underline STFPM’s potential for scalable, automated non-destructive testing in real-world industrial applications.

Future work should focus on refining localization accuracy and adapting models to varying defect types and materials to support broader deployment of automated shearographic inspection.

\begin{credits}
\subsubsection{\ackname} This research was funded by the Ministry of Science and Health Rhineland-Palatinate as part of the Young Researchers Fund of the Research Initiative (Funding Period 2025) and by the Federal Ministry for Economic Affairs and Climate Protection (BMWK) through the ‘Central Innovation Program for SMEs (ZIM)’, funding code KK5060010SY3.

\subsubsection{\discintname}
The authors declare no conflict of interest.
\end{credits}

\end{document}